\documentclass[sigconf]{acmart}

\usepackage{framed}
\newcommand{\highlight}[1]{\begin{framed}%
  \noindent\emph{#1}
\end{framed}}

\AtBeginDocument{%
  }

\acmConference[ICSE 2024]{46th International Conference on Software
  Engineering}{April 2024}{Lisbon, Portugal}

\begin{document}

\title{Data vs. Model Machine Learning Fairness Testing: An Empirical Study}

\author{Arumoy Shome}
\affiliation{%
  \institution{Delft University of Technology}
  \city{Delft}
  \country{Netherlands}}
\email{a.shome@tudelft.nl}

\author{Lu{\'\i}s Cruz}
\affiliation{%
  \institution{Delft University of Technology}
  \city{Delft}
  \country{Netherlands}}
\email{l.cruz@tudelft.nl}

\author{Arie van Deursen}
\affiliation{%
 \institution{Delft University of Technology}
 \city{Delft}
\country{Netherlands}}
\email{arie.vandeursen@tudelft.nl}

\begin{abstract}

Although several fairness definitions and bias mitigation techniques
  exist in the literature, all existing solutions evaluate fairness of
  Machine Learning (ML) systems after the training stage. In this
  paper, we take the first steps towards evaluating a more holistic
  approach by testing for fairness both before and after model
  training. We evaluate the effectiveness of the proposed approach and
  position it within the ML development lifecycle, using an empirical
  analysis of the relationship between model dependent and independent
  fairness metrics. The study uses 2 fairness metrics, 4 ML
  algorithms, 5 real-world datasets and 1600 fairness evaluation
  cycles. We find a linear relationship between data and model
  fairness metrics when the distribution and the size of the training
  data changes. Our results indicate that testing for fairness prior
  to training can be a ``cheap'' and effective means of catching
  a biased data collection process early; detecting data drifts in
  production systems and minimising execution of full training cycles
  thus reducing development time and costs.

\end{abstract}

\maketitle

\section{Introduction}\label{sec:intro}

While several contributions toward testing ML systems have been made
in recent years, preference has primarily been given to robustness and
correctness while other non-functional properties such as security,
privacy, efficiency, interpretability and fairness have been
ignored \cite{zhang2020machine,zhang2021ignorance,mehrabi2021survey,wan2021modeling}.
Testing for fairness in ML systems however, is a multi-faceted problem
and involves both technological and social factors. Although an
abundance of definitions for fairness and consequently techniques to
mitigate said bias exists in the scientific literature, all existing
solutions evaluate fairness after the training stage, using the
predictions of the ML model.

In contrast to prior work, we take a more holistic approach by testing
for fairness at two distinct locations of the ML development
lifecycle. First, prior to model training using fairness metrics that
can quantify the bias in the training data (henceforth Data Fairness
Metric or DFM). And second, after model training using fairness
metrics that quantify the bias in the predictions of the trained model
(henceforth Model Fairness Metric or MFM).

While MFM has been widely adopted in practice and well researched in
academia, we do not yet know the role of DFM when testing for fairness
in ML systems. The research goal of this study is to evaluate the
effectiveness of DFM for catching fairness bugs. We do this by
analysing the relationship between DFM and MFM through an extensive
empirical study. The analysis is conducted using $2$ fairness metrics,
$4$ ML algorithms, $5$ real-world tabular datasets and $1600$ fairness
evaluation cycles. To the best of our knowledge, this is the first
study which attempts to bridge this gap in scientific knowledge. Our
results are exploratory and open several intriguing avenues of
research.

The research questions along with the contributions of this paper are
as follows.
\begin{enumerate}
  \item[RQ1.] \textbf{What is the relationship between DFM and MFM as the
    fairness properties of the underlying training dataset change?}

    DFM and MFM convey the same information when the distribution of
    the underlying training dataset changes. This implies that DFM can
    be used as early warning systems to catch data drifts in
    production ML systems that may affect its fairness.

  \item[RQ2.] \textbf{How does the training sample size affect the
    relationship between DFM and MFM?}

    Our analysis of the training sample size and how it influences the
    relationship between DFM and MFM reveals the presence of a
    trade-off between fairness, efficiency and correctness. In
    Section \ref{sec:discuss-fair-eff-perf-trade} we provide some
    practical guidelines on how to best navigate this trade-off.

  \item[RQ3.] \textbf{What is the relationship between DFM and MFM across
    various training and feature sample sizes?}

    DFM and MFM convey the same information when the training sample
    size changes. This implies that DFM can help practitioners catch
    fairness issues upstream and avoid execution costs of a full
    training cycle.
\end{enumerate}

All source code and results of the study are publicly accessible under the CC-BY 4.0 license\footnote{The replication package of this study is available here: https://figshare.com/s/67206f7c219b12885a6f}.
\section{Preliminaries}\label{sec:related}

\subsection{Algorithmic Bias, Bias Mitigation and Group Fairness}\label{sec:bias-fairness}

Several fairness metrics have been developed to mitigate the costs and
human biases when manually validating the fairness of the labels in
the dataset. Fairness metrics primarily focus on supervised binary
classification with protected attributes like \emph{race} or
\emph{sex}. They can be broadly classified into group fairness
(ensuring similar predictions across different groups) and individual
fairness (consistent predictions for individuals differing only in
protected
attributes)~\cite{mitchell2021algorithmic,barocas2019fairness,hardt2016equality}.
Addressing bias requires appropriate mitigation techniques, which can
be categorised into pre-processing (altering training
data)~\cite{feldman2015certifying,zemel2013learning}, in-processing
(integrating fairness in model
training)~\cite{zhang2018mitigating,agarwal2018reductions,kearns2018preventing},
and post-processing (adjusting model
predictions)~\cite{pleiss2017fairness,hardt2016equality,kamiran2012decision}
techniques. Each category aims to ensure fairness in ML models, either
by adjusting data, the training process, or the outcomes.

This study uses group fairness metrics due to their popularity in
existing empirical studies on ML fairness testing and ease of
understandability \cite{zhang2021ignorance,biswas2020machine,biswas2021fair,hort2021fairea}.
Our analysis of the relationship between the DFM and MFM (see
Section \ref{sec:implications}) presents practical guidelines for
practitioners to pick the appropriate bias mitigation strategy based
on the particular fairness issue they are facing.
\subsection{Prior Work in ML Fairness Testing}\label{sec:prior-work}

\begin{figure*}
  \centering
  \includegraphics[width=0.8\linewidth]{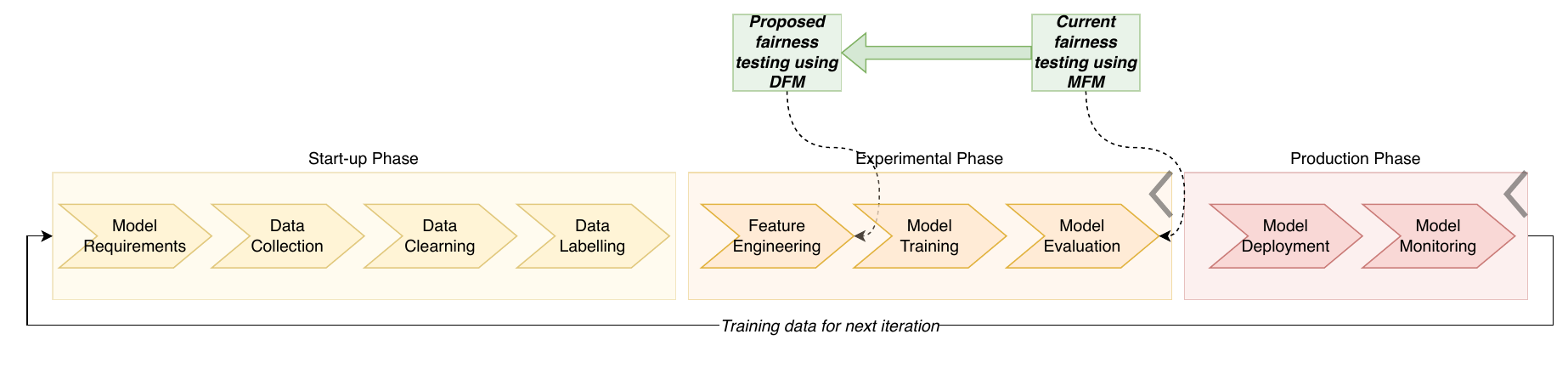}
  \caption{Stages of the ML Lifecycle (adopted from
    \cite{amershi2019software,breck2019data}). Three
    distinct phases of the lifecycle are marked by different colours.
    Stages in the experimental and production phases may loop back to
    any prior stages, indicated by the large grey arrows. The location
    of fairness testing using DFM and MFM are marked by the green
    labels. The green arrow depicts the shift proposed by this study
    in ML fairness testing. }
  \label{fig:ml-lifecycle}
\end{figure*}

Recent literature has extensively reviewed ML fairness testing and
bias mitigation methods. \emph{Wan et al.
(2021)}~\cite{wan2021modeling} focused on in-processing bias
mitigation, while \emph{Chen et al. (2022)}~\cite{chen2022fairness}
and \emph{Mehrabi et al. (2021)}~\cite{mehrabi2021survey} surveyed
fairness testing in ML, with the latter emphasizing fairness issues
related to both data and models. \emph{Chen et al. (2022)} also
defined fairness bugs and testing in ML from a software engineering
perspective, differentiating it from traditional software testing.
\emph{Biswas et al. (2021)}~\cite{biswas2021fair} analyzed the impact
of common data pre-processing techniques on ML model fairness, using
real-world ML pipelines. \emph{Feffer et al.
(2022)}~\cite{feffer2022empirical} examined the effectiveness of
combining bias mitigation with ensemble techniques, and \emph{Zhang et
al. (2021)}~\cite{zhang2021ignorance} studied how training sample size
and feature sample size affect model fairness.

In contrast to white-box testing, where the training data and
algorithm are known, black-box testing treats the ML pipeline as
opaque. This approach has led to the development of test input
generation tools like Themis by \emph{Galhotra et al.
(2017)}~\cite{galhotra2017fairness} for causal fairness testing,
Aequitas by \emph{Udeshi et al. (2018)}~\cite{udeshi2018automated} for
detecting discriminatory model behavior, and a technique by
\emph{Aggarwal et al. (2019)}~\cite{aggarwal2019black} using symbolic
execution to consider correlations between protected and unprotected
attributes.

This study conducts an empirical analysis of the relationship between
DFM and MFM, and as such, operates under white-box testing
assumptions. The experimental design of this study is similar in
spirit to that proposed by \emph{Zhang et al.
(2021)}~\cite{zhang2021ignorance}. However our objective, results and
implications are entirely different. While \emph{Zhang et al.
(2021)}~\cite{zhang2021ignorance} study the effect of training and
feature sample size on the fairness of the model, this study aims to
understand the relationship between DFM and MFM. We analyse how change
in the distribution, sample size and number of features in the
training set affects this relationship. Through this analysis, we hope
to establish the role of DFM within the existing ML development
lifecycle.
\subsection{Machine Learning Lifecycle and Fairness
  Testing}\label{sec:ml-lifecycle}

Figure \ref{fig:ml-lifecycle} illustrates the ML lifecycle, dividing
it into three phases with feedback loops allowing stages to influence
each other, particularly in the experimental and production phases.
Initially, the start-up phase is data-centric, involving data
collection, cleaning, and labeling, especially for safety-critical
applications. The subsequent experimental phase focuses on feature
engineering and optimizing ML model performance, incorporating not
only functional properties like correctness but also non-functional
aspects like fairness and security. In the production phase, the ML
model---unlike traditional software---continues operating despite
errors, necessitating continuous monitoring and updates when
performance falls below a certain
threshold~\cite{amershi2019software,breck2019data,sambasivan2021everyone,zhang2020machine}.

This study explores the implications of assessing ML model fairness
before training, marking a shift from the traditional post-training
fairness evaluation.
\section{Experimental Design}\label{sec:method}

\begin{figure*}
  \centering
  \includegraphics[width=0.8\linewidth]{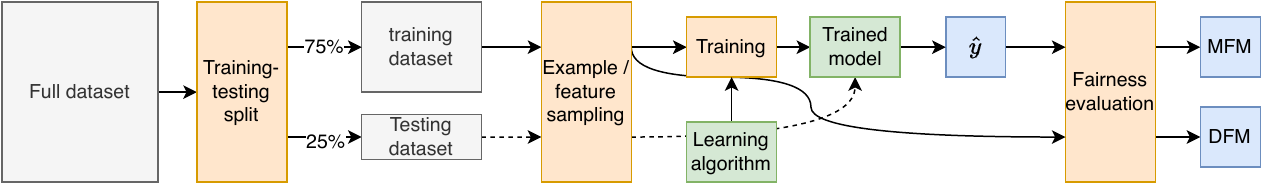}
  \caption{Methodology for evaluating fairness of datasets and ML
  models using DFM and MFM.}
  \label{fig:method}
\end{figure*}

\subsection{Datasets, ML Models and Fairness Metrics}\label{sec:method-parameters}

\begin{table}
  \centering
  \caption{Fairness metrics used in the study}
  \begin{tabular}{l r}
    \hline
    \textbf{\emph{DFM}}\\
    \hline
    DI & \(\displaystyle \frac{P(Y=1|D=0)}{P(Y=1|D=1)}\)\\
    SPD & \(\displaystyle P(Y=1|D=0)-P(Y=1|D=1)\)\\
    \hline
    \textbf{\emph{MFM}}\\
    \hline
    DI & \(\displaystyle \frac{P(\hat{Y}=1|D=0)}{P(\hat{Y}=1|D=1)}\)\\
    SPD & \(\displaystyle P(\hat{Y}=1|D=0)-P(\hat{Y}=1|D=1)\)\\
    \hline
  \end{tabular}
  \label{tab:fairness-metrics}
\end{table}

Table \ref{tab:fairness-metrics} shows the group fairness metrics
along with their mathematical formulas used in this study. We include
all group fairness metrics---namely \emph{Disparate Impact (DI)} and
\emph{Statistical Parity Difference (SPD)}---for which both model
dependent and independent variants are available. The DFM use the
labels of the data ($Y$) whereas the MFM use the predictions of the
trained ML models ($\hat{Y}$). Favourable and unfavourable outcomes
are represented by $1$ and $0$ respectively. Similarly, privileged and
unprivileged groups of the protected attribute ($D$) are represented
by $1$ and $0$ respectively. All fairness metrics and datasets used in
this study are obtained from the \emph{AIF360} python
library \cite{bellamy2019ai}.

\begin{table}
  \centering
  \caption{Datasets used in the study}
  \begin{tabular}{l l r}
    \hline
    \textbf{Name} & \textbf{Prot.} & \textbf{\#Eg.}\\
    \hline
    German \cite{hofmann1994german} & age, sex & 1000\\
    Compas\cite{angwin2016machine} & race, sex & 6167\\
    MEPS \cite{mepsdata} & race & 15675\\
    Bank\cite{moro2014data} & age & 30488\\
    Adult\cite{kohavi1996scaling} & race, sex & 45222\\
    \hline
  \end{tabular}
  \label{tab:datasets}
\end{table}

Table \ref{tab:datasets} presents the datasets used in this study. We
consider tabular datasets which have been extensively used in prior
scientific contributions on ML fairness
testing \cite{zhang2021ignorance,biswas2020machine,biswas2021fair,chen2022fairness}. Based
on prior work, we only consider one protected attribute at any given
time thus giving us eight independent datasets. We follow the default
pre-processing steps implemented in the AIF360 library---missing
values are dropped and categorical features are label encoded. Prior
to training, the features in the training and testing subsets are
standardised by removing the mean of the sample and scaling to unit
variance.

We use the scikit-learn \cite{pedregosa2011scikit} python library for
creating the train-test splits and training the ML models. We use four
ML models of varying complexity namely, \emph{Logistic Regression},
\emph{Decision Trees}, \emph{Random Forest} and \emph{Ada boost} based
on their popularity in practice and in prior scientific
publications \cite{zhang2021ignorance,biswas2021fair,biswas2020machine}.
\subsection{Fairness Evaluation}\label{sec:method-fair-eval}

\begin{table}
  \centering
  \caption{Parameters of the study}
  \begin{tabular}{l r}
    \hline
    \textbf{Parameter} & \textbf{Count}\\
    \hline
    Fairness metrics & $2$\\
    ML models & $4$\\
    Datasets & $8$\\
    Total cases & $8\times4=32$\\
    Iterations & $50$\\
    Total fairness evaluation cycles & $32\times50=1600$\\
    \hline
  \end{tabular}
  \label{tab:parameters}
\end{table}

Figure \ref{fig:method} presents the methodology used in this study
for evaluating the fairness of ML models and datasets. A 75--25 split
with shuffling is used to create the training and testing splits. DFMs
and MFMs are used to quantify the bias in the underlying distribution
of the training set and the predictions of the models respectively. We
adopted the transformation steps from prior work to scale all fairness
metric values between $0$ and $1$ such that higher values indicate
more bias \cite{zhang2021ignorance,hort2021fairea}.

We extend the above experiment further in two ways. First, we
experiment with different number of examples and second with different
number of features in the training set. For both experiments, we
shuffle the order of the examples in the training and testing sets.
Additionally, for the feature sample size experiment we shuffle the
order of the features.

For the training sample size experiment, we generate different
training samples of varying sizes starting from 10\% of the original
training data, and increase in steps of 10\% until the full quantity
is reached. For the feature sample size experiment, we start with
a minimum of three features (in addition to the protected attribute
and target) and introduce one new feature until all the features are
utilised. Both the training and testing sets undergo the same feature
sampling procedure in the feature sample size experiment. No such
sampling is done in the testing set for the training sample size
experiment.

Table \ref{tab:parameters} summarises the parameters of the study. We
train 4 ML models on 8 datasets producing 32 total cases. The fairness
for each case is evaluated 50 times using two fairness metrics, thus
producing a total of 1600 training and fairness evaluation cycles.
\subsection{Correlation Analysis}\label{sec:corr-analysis}

We use correlation analysis to study the relationship between DFM and
MFM with-respect-to change in three experimental
factors---distribution, size and features of the training set.
Spearman Rank Correlation is used to quantify the linear relationship
between the DFM and MFM since it does not assume normality and is
robust to outliers. We repeat all experiments 50 times and report the
statistical significance of our results. We consider cases where
$pvalue\le0.05$ to be statistically significant in our evaluation.

We do not apply the \emph{Bonferroni correction} to the correlation
analysis results. Although we report the $pvalue$ for completeness, we
do not base our implications only on the statistically significant
results. But rather on general trends observed in our analysis.

\begin{table}
\centering
\caption{Interpretation of correlation analysis used in this study.}
\begin{tabular}{p{0.1\linewidth} p{0.15\linewidth} p{0.15\linewidth} p{0.4\linewidth}}
\hline
\textbf{Corr.} &
\textbf{Direction} &
\textbf{Fairness} &
\textbf{Cause}\\
\hline
+ &
Same &
Same info. &
DFM and MFM both report high/low values\\
  0 &
  NA &
  NA &
  No bias in training data or DFM and MFM do not report similarly\\
  - &
  Opposite &
  Not same info. &
  MFM lower than DFM (opposite is not possible)\\
\hline
\end{tabular}
\label{tab:corr-summary}
\end{table}

Table~\ref{tab:corr-summary} summaries the results of the correlation
analysis along with our interpretation. A positive correlation means
that the DFM and MFM changed in the same direction and thus \emph{convey
the same information}. A positive correlation indicates that both the
DFM and MFM show presence of bias in the training data and in the model
predictions respectively.

A negative correlation between the DFM and MFM means that they changed
in the opposite direction and \emph{do not convey the same
information}. This indicates that the bias in the model predictions
was lower than that present in the training data. Note that the
opposite event---of bias in the model predictions being higher than
that present in the training data---is unlikely to occur. ML models
cannot manifest bias towards a particular protected group if the
training data is unbiased to begin with.

The implications of the correlation results varies based on the
experimental factor. A detailed discussion is presented in the
corresponding sub-sections of Section \ref{sec:implications}.
\subsection{Distribution of Experimental Data}\label{sec:data-analysis}

\begin{figure}
  \centering
  \includegraphics[width=\linewidth]{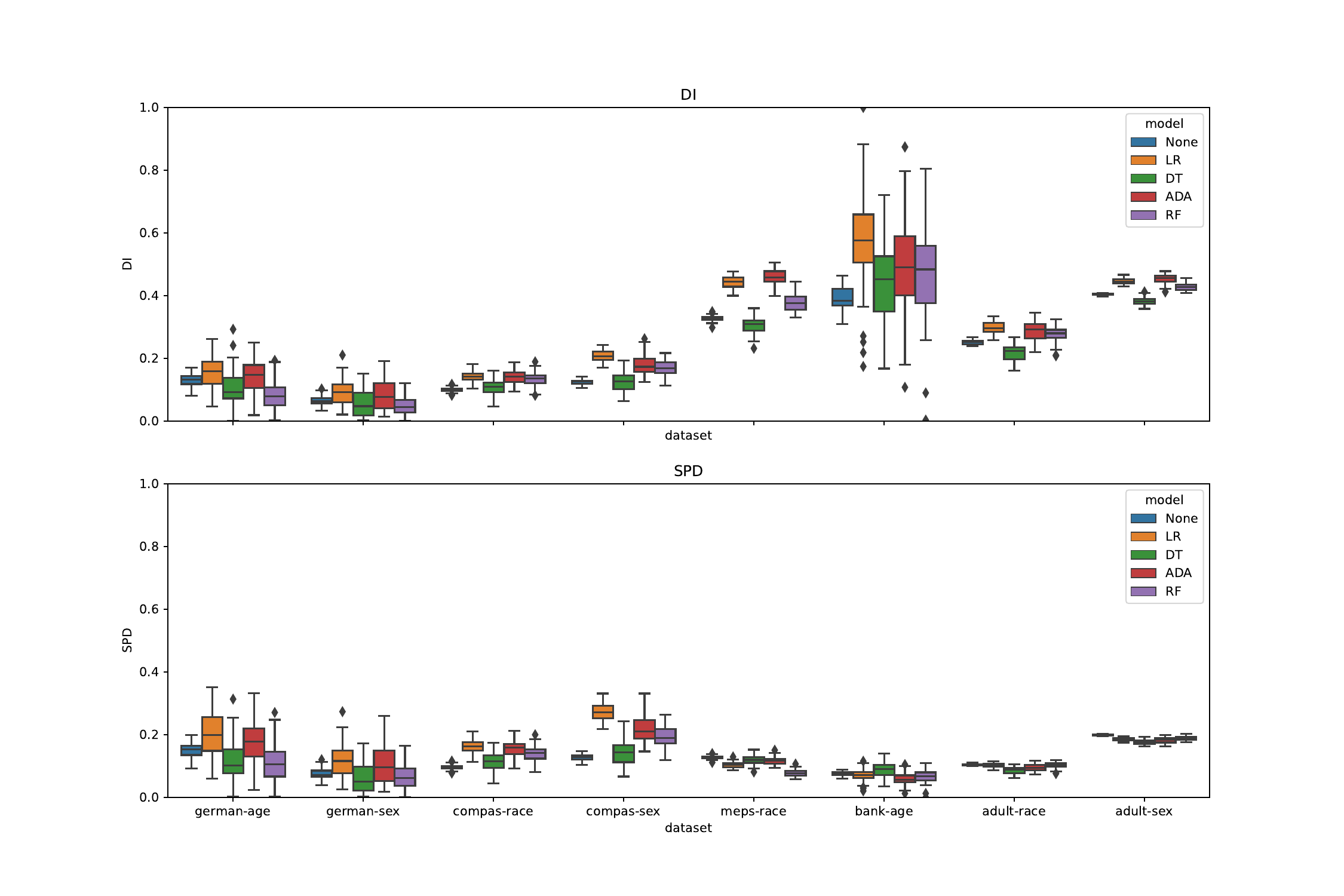}
  \caption{Boxplot showing distribution of DFM and MFM for all
    datasets, models and fairness metrics.}
  \label{fig:boxplot--dataset--di-spd--exp-full}
\end{figure}

Figure \ref{fig:boxplot--dataset--di-spd--exp-full} presents a boxplot
with the distribution of DFM and MFM for all datasets, models and
fairness metrics. The x-axis represents the datasets used in this
study while the y-axis presents the value of the fairness metric. The
models used in this study are represented using different
colours---note that the model ``None'' represented in blue refers to
the DFM. Both the fairness metrics DI (top) and SPD (bottom) are
presented in separate plots.

Comparing the boxplots for DI and SDP, we note that they follow a
similar pattern of distribution. Comparing the DFM and MFM for any
given ML model and dataset, we observe a difference in the
distribution of the DFM and MFM. In several instances the tree-based
classifiers (DT and RF) make fairer decisions compared to the other
classifiers, sometimes even better than the baseline provided by the
DFM.

The variability of MFM tends to be higher than DFM. This is because,
in addition to the randomness from the data shuffling in the training
set, the models are assigned random initial states in every iteration
thus resulting in different predictions across the iterations and
consequently varying MFM values.
\section{Results}\label{sec:results}

\subsection{RQ1. What is the relationship between DFM and MFM as
the fairness properties of the underlying training dataset
changes?}\label{sec:results-full-rel}

\begin{figure}
  \centering
  \includegraphics[width=\linewidth]{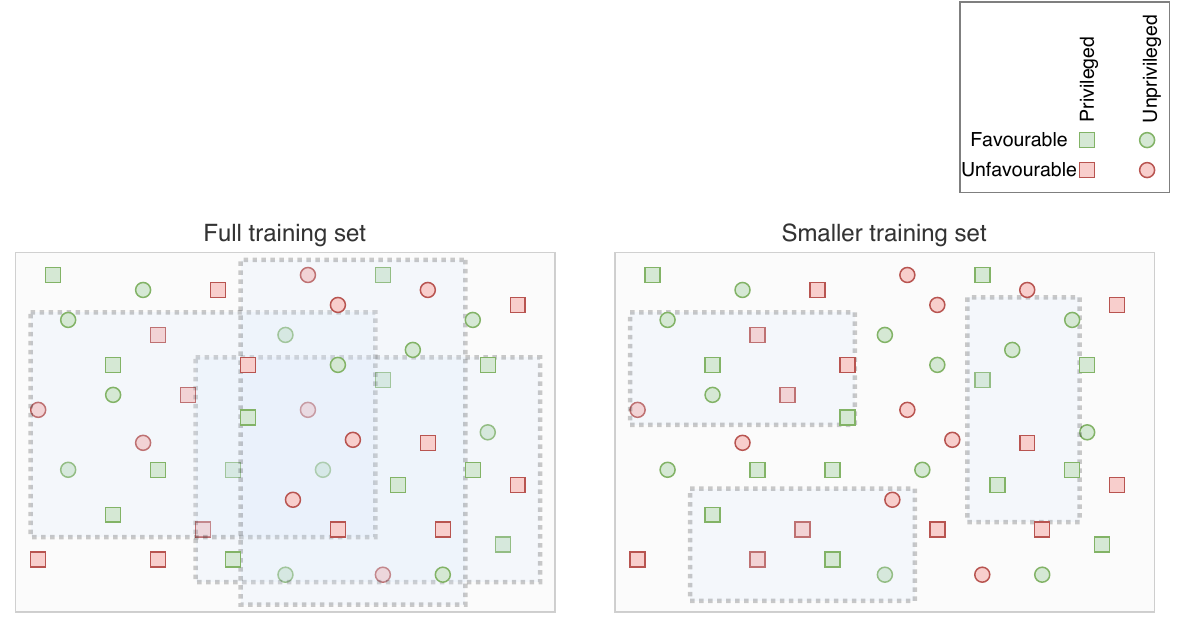}
  \caption{Visual explanation of rationale for using smaller training
    sample to simulate change in the distribution of the training
    data. The grey boxes represent the full dataset while the blue
    boxes represent the training set for three hypothetical
    iterations. More overlap in the blue boxes depicts less
    distribution change and vice-versa.}
  \label{fig:shuffle}
\end{figure}

\begin{figure}
  \centering
  \includegraphics[width=\linewidth]{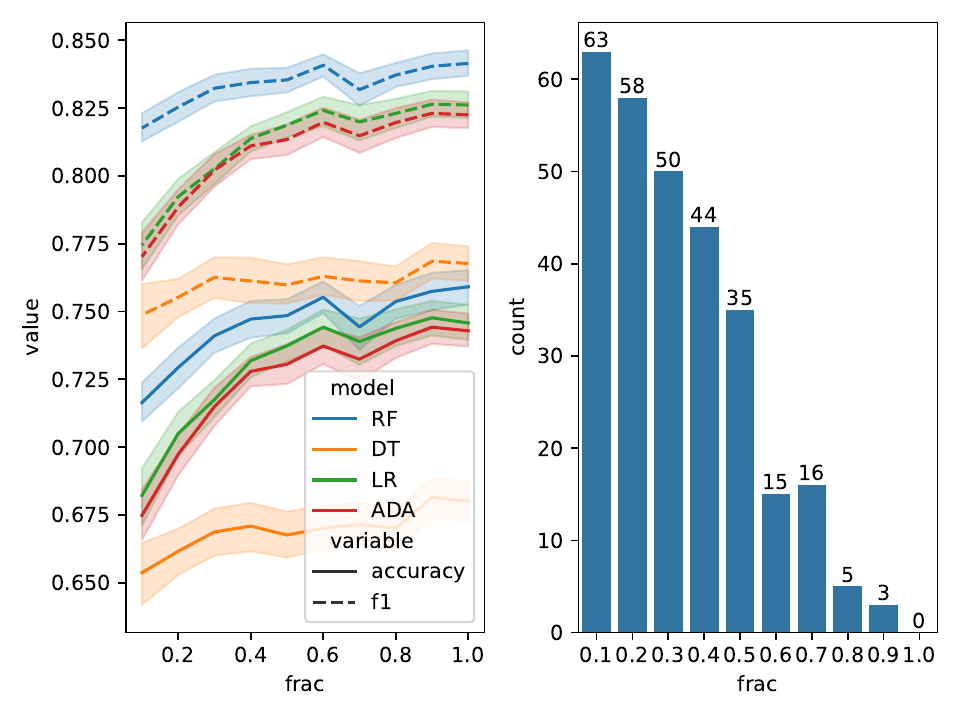}
  \caption{\emph{\textbf{(left)}} Lineplot showing relationship
    between performance metrics and training sample size in the
    \emph{german-age} dataset. Data from the 50 iterations is
    aggregated using the mean, the error bars show the standard
    deviation. \emph{\textbf{(right)}} Countplot showing number of
    cases with significant change in accuracy and f1 when trained
    using the full vs. smaller training sample size.}
  \label{fig:training-set-frac-threshold}
\end{figure}

\begin{figure}
  \centering
  \includegraphics[width=\linewidth]{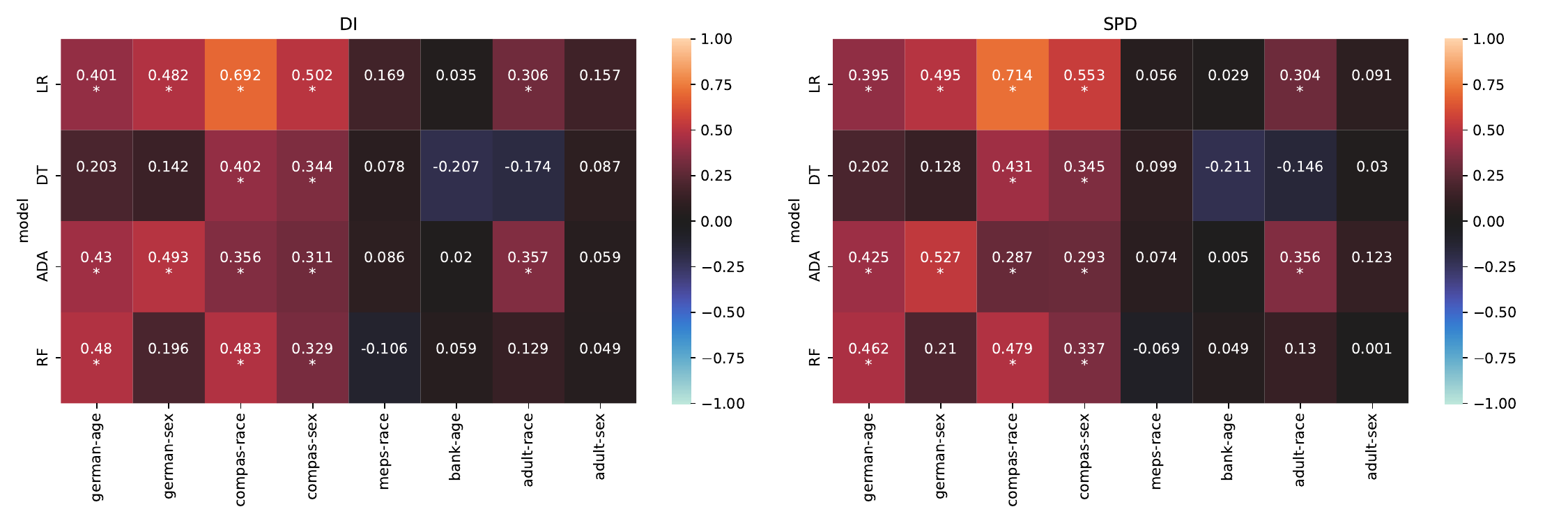}
  \caption{Heatmap showing correlation between DFM and MFM for all
    models and datasets using training data with simulated
    distribution change. Each block is representative of 50
    iterations. The statistically significant cases are marked with an
    asterisks. We primarily observe bright hues of red indicating that
    the DFM and MFM convey the same information. This means that DFM
    can be used to identify fairness related data drifts in automated
    ML pipelines.}
  \label{fig:heatmap--corr--training-sets-frac}
\end{figure}

\begin{figure}
  \centering
  \includegraphics[width=\linewidth]{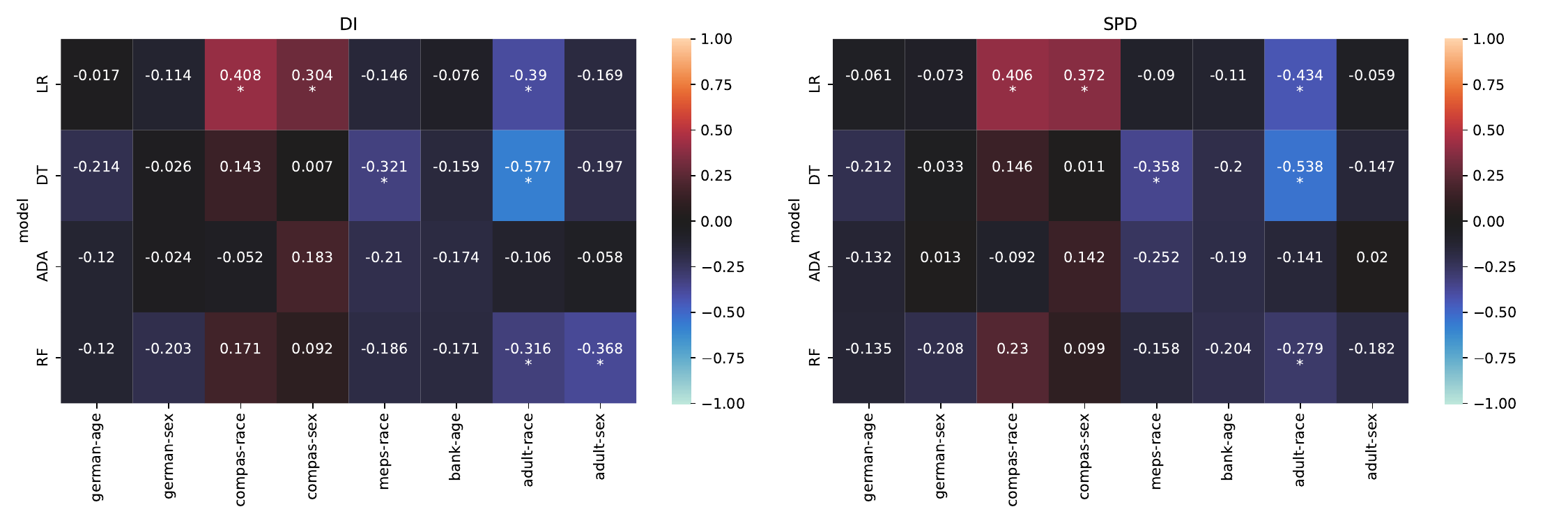}
  \caption{Heatmap showing correlation between DFM and MFM using
    training data without significant distribution change. Each block
    is representative of 50 iterations. In contrast to
    Figure \ref{fig:heatmap--corr--training-sets-frac}, we primarily
    observe darker colours indicating that the DFM and MFM no longer
    convey the same information.}
  \label{fig:heatmap--corr--full-data}
\end{figure}

To simulate change in the distribution of the training data, we use a
smaller subset of the original training data. Figure \ref{fig:shuffle}
visually motivates this approach. Privileged examples ($D=1$) are
represented using squares while unprivileged examples ($D=0$) are
represented using circular points. The label of the examples are
represented using colour---green for favourable ($Y=1$) and red for
unfavourable ($Y=0$) examples. The large grey boxes represent the
original dataset while the blue boxes represent the training dataset.

The figures present three hypothetical iterations of the data sampling
process to create the training set. The left figure presents the
scenario where the full training set is used while the right figure
presents the scenario where a smaller portion of the full training set
is used. As seen from the figure, there is more overlap amongst the
examples in the full training set in every iteration. In contrast,
there is less overlap amongst the examples in the smaller training
set. The distribution in smaller training samples will thus change
more frequently in the 50 iterations and capture a wide variety of
data fairness properties.

The data quality in smaller training samples will however deteriorate.
To identify a sample size that captures a variety of data fairness
properties while also being a realistic training dataset, we analyse
the \emph{accuracy} and \emph{f1 score} of the models across different
training sample sizes (as explained in
Section \ref{sec:method-fair-eval}). Next, we conduct \emph{student
t-test} to identify the smallest sample size where the performance of
the models is similar to that obtained when trained using the full
training set.

Figure \ref{fig:training-set-frac-threshold} (right) presents a
histogram of the number of cases where there was a significant
difference between the two populations. We note that there is a
significant difference in the performance of the models, in the
majority of the cases when the training size is reduced to 50\%. The
performance however remains consistent when using a training size of
60\% or higher. This is also corroborated by the lineplot in
Figure \ref{fig:training-set-frac-threshold} (left) which shows the
accuracy and f1 of all models across various training sample sizes in
the \emph{german-age} dataset. We observe that the performance
stabilises starting from 60\% training sample size. Thus for the
majority of the cases, a training sample of 60\% allows us to train
models with acceptable performance, while also capturing a wide
variety of fairness issues in the underlying training data within the
50 iterations.

Figure \ref{fig:heatmap--corr--training-sets-frac} shows the
correlation between the DFM and MFM across all models and datasets
when trained using data with simulated distribution change as outlined
above. The models used in this study are represented along the y-axis
and the datasets along the x-axis. Darker colours indicate weaker
correlation whereas brighter colours indicate stronger correlation.
Positive correlation is indicated using hues of red while negative
correlation is indicated using hues of blue. The correlation between
DFM and MFM for both fairness metrics are shown separately. We
primarily observe a positive correlation between the DFM and MFM. This
indicates that the DFM and the MFM convey the same information as the
distribution---and consequently the fairness properties---of the
underlying training dataset changes.

In contrast to Figure \ref{fig:heatmap--corr--training-sets-frac},
Figure \ref{fig:heatmap--corr--full-data} shows the correlation
between DFM and MFM when trained using data without sufficient
distribution change. Due to lack of significant change in the
distribution of the training data, we primarily observe darker colours
indicating that the DFM and MFM are not linearly related to one
another anymore.

\highlight{\textbf{Answer to RQ1:} DFM and MFM are positively
correlated and thus convey the same information as the
distribution---and consequently the fairness properties---of the
underlying training dataset changes.}

\subsection{RQ2. How does the training sample size affect the
correlation between DFM and MFM?}\label{sec:results-corr-frac}

\begin{figure}
  \centering
  \includegraphics[width=0.8\linewidth]{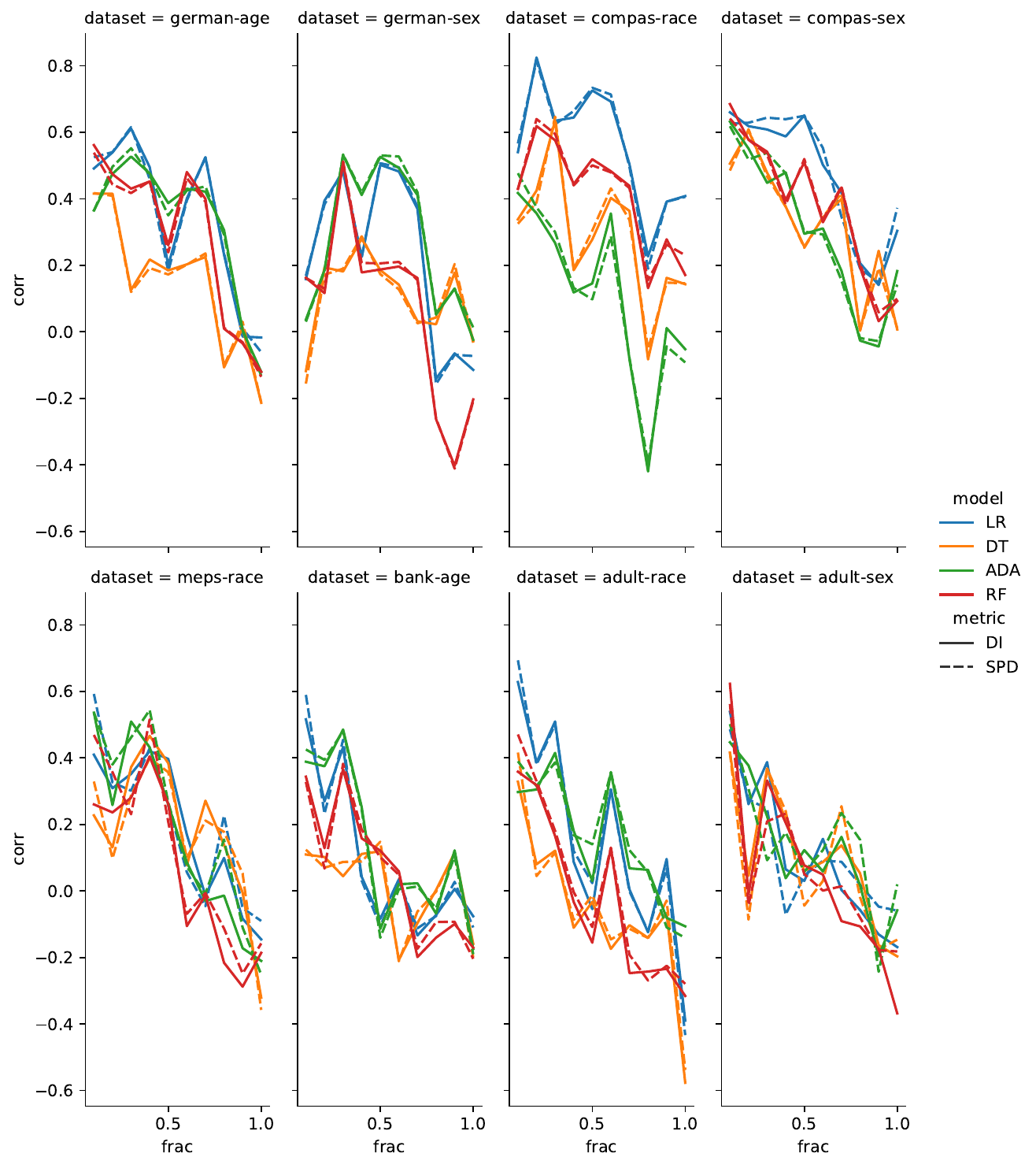}
  \caption{Lineplot showing relationship of correlation (between DFM
    and MFM) with training sample size for all datasets and models.
    Data from the 50 iterations is aggregated using the mean value.
    The correlation decreases as the training sample size is
    increased. This indicates that with sufficient training data, ML
    models are able to circumvent some of the bias in the underlying
    training set.}
  \label{fig:lineplot--frac--corr}
\end{figure}

In Figure \ref{fig:heatmap--corr--training-sets-frac}, the correlation
in the smaller datasets is more positive compared to the larger
datasets when trained using a smaller training sample size. When the
training sample size is increased, the correlation in the datasets
decrease as observed in Figure \ref{fig:heatmap--corr--full-data}.

Based on these observations, we hypothesise that the quantity of
training data influences the correlation between DFM and MFM. Our
hypothesis is corroborated by Figure \ref{fig:lineplot--frac--corr}
which shows the relationship of the correlation with various training
sample sizes, for all datasets and models. The x-axis presents the
training sample size and the y-axis presents the correlation between
the DFM and MFM. The colours represent the models while the style of
the line represents the two fairness metrics. Each dataset is shown as
a separate subplot. The overwhelming majority show that the
correlation between the DFM and MFM decreases as we increase the
training sample size.

\highlight{\textbf{Answer to RQ2:} The training sample size has
a profound effect on the relationship between the DFM and MFM. The
correlation between the DFM and MFM decreases as we increase the
training sample size.}

\subsection{RQ3. What is the relationship between DFM and MFM
across various training and feature sample
sizes?}\label{sec:results-size-feature-sample}

\begin{figure}
  \centering
  \includegraphics[width=\linewidth]{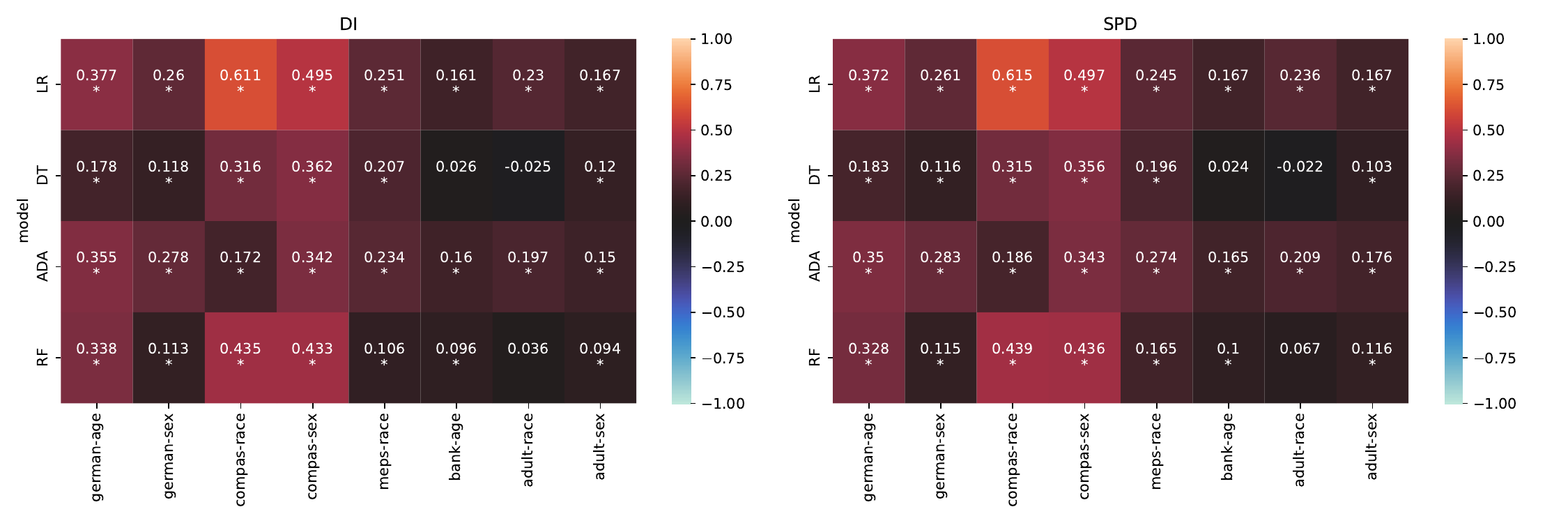}
  \caption{Heatmap showing correlation between DFM and MFM across all
    training sample sizes. Each block is representative of 50
    iterations $\times$ 10 training sample size $=$ 500 data points.
    The statistically significant cases are marked with an asterisks.
    We primarily observe hues of red indicating that the DFM and MFM
    convey the same information. When experimenting with the training
    sample size, DFM can aid practitioners catch fairness issues
    upstream and avoid execution costs of a full training cycle.}
  \label{fig:heatmap--corr--frac}
\end{figure}

\begin{figure}
  \centering
  \includegraphics[width=\linewidth]{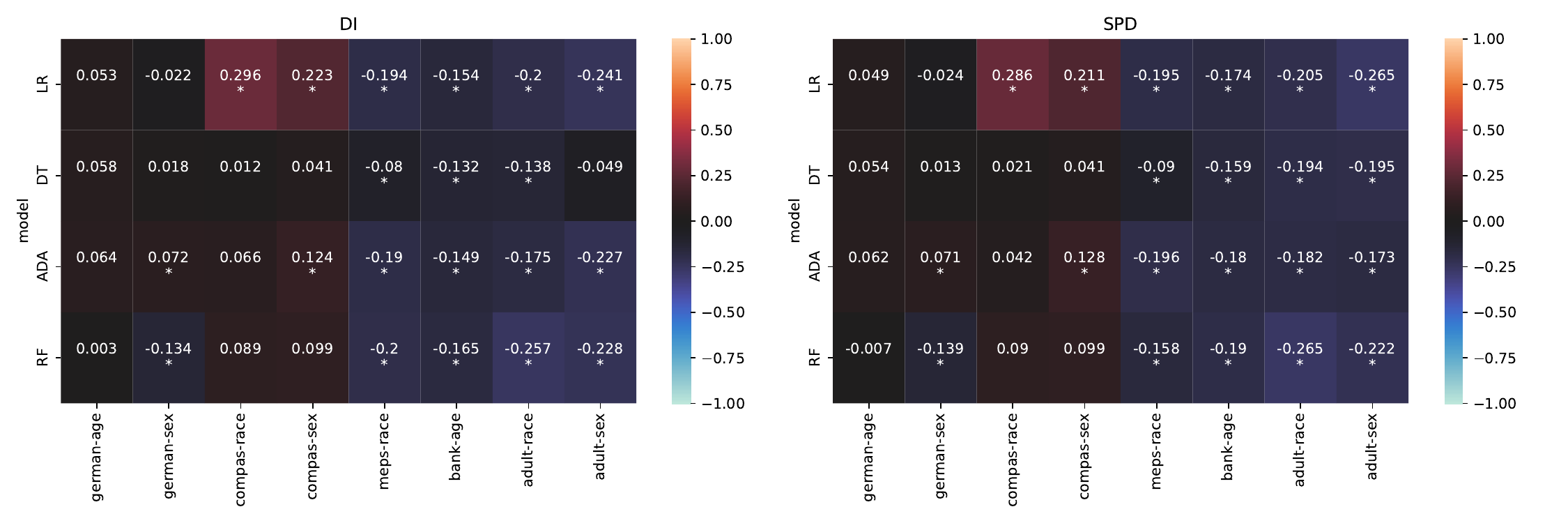}
  \caption{Heatmap showing correlation between DFM and MFM across
    various feature sample sizes. Each block is representative of data
    points from 50 iterations $\times$ the number of features in the
    dataset. We primarily observe darker colours indicating that the
    DFM and MFM do not convey the same information. When experimenting
    with the feature sample size, practitioners must test for fairness
    both before and after training.}

  \label{fig:heatmap--corr--num-features}
\end{figure}

\begin{figure}
  \centering
  \includegraphics[width=0.8\linewidth]{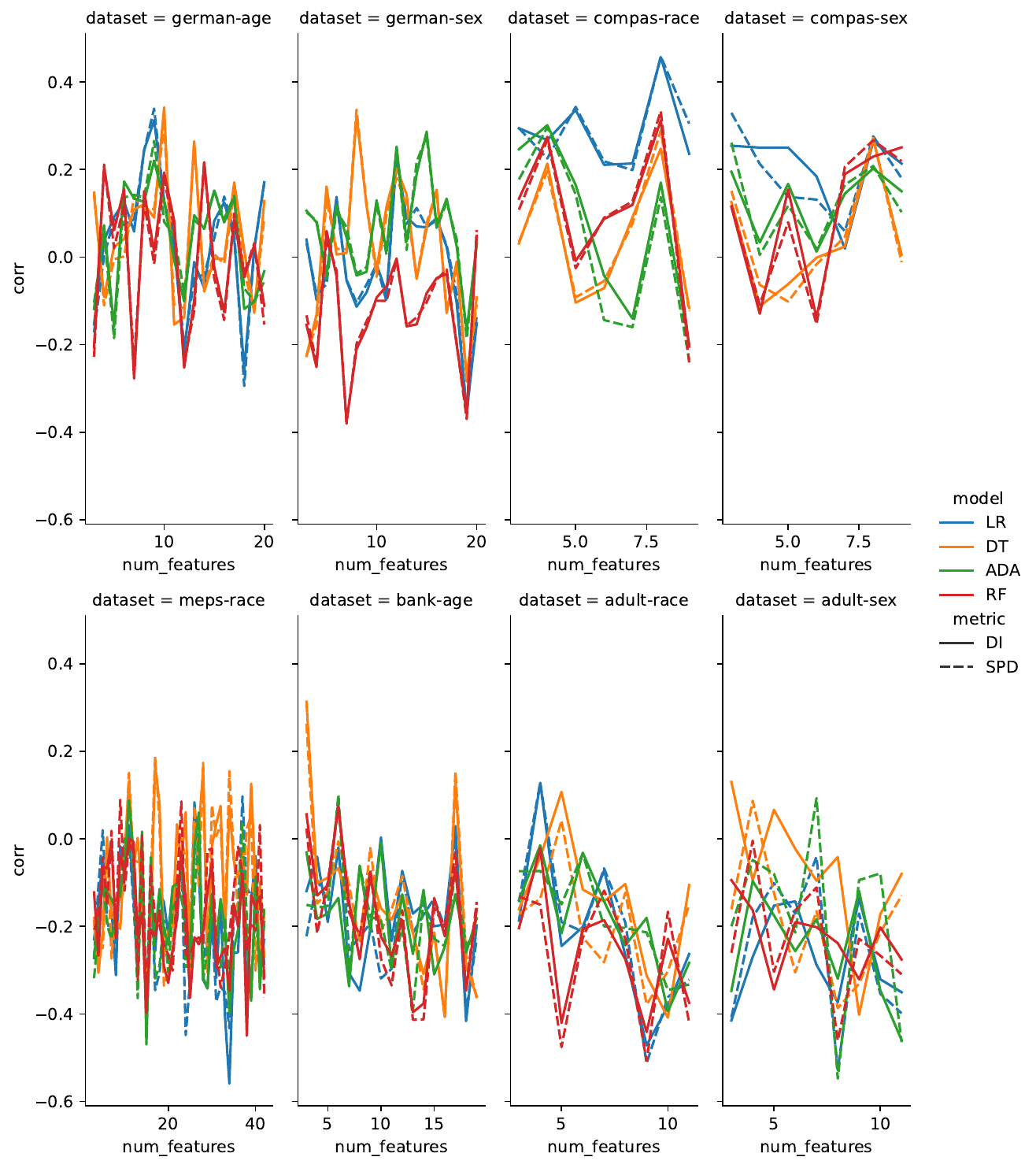}
  \caption{Lineplot showing relationship of correlation (between DFM
    and MFM) with feature sample size for all datasets and models.
    Data from the 50 iterations is aggregated using the mean value.
    There is no discernible relationship since the number of features
    does not influence the DFM.}
  \label{fig:lineplot--num-features--corr}
\end{figure}

\subsubsection{Training Sample Size}

In this section we analyse the relationship between DFM and MFM across
varying training sample sizes. In contrast to
Section \ref{sec:results-corr-frac} where we calculated the
correlation between DFM and MFM within each training sample size, here
we calculate the correlation across all training sample sizes. The
correlation between the DFM and MFM is shown in
Figure \ref{fig:heatmap--corr--frac}. We primarily observe colours
indicating that the DFM and MFM convey similar information as the
training sample size changes.

\subsubsection{Feature Sample Size}

In this section we analyse the relationship between the DFM and MFM
across varying feature sample sizes. In contrast to the training
sample size experiment above, we change the number of features in the
training set and randomise the feature order in each iteration.
Figure \ref{fig:heatmap--corr--num-features} presents the correlation
between the DFM and MFM across all feature sample sizes. We primarily
notice darker colours indicating that there is no significant
correlation between the DFM and MFM as the number of features in the
training dataset changes.

From Table \ref{tab:fairness-metrics}, we note that the feature sample
size does not affect the DFM thus explaining the lack of significant
correlation between the DFM and MFM. The larger datasets show a more
negative correlation. As explained in
Section \ref{sec:results-corr-frac}, this is because the larger
datasets have sufficient training data for the model to mitigate some
of the bias in the underlying training set. This can also be verified
in Figure \ref{fig:lineplot--num-features--corr} which shows the
relationship between the correlation and the feature sample sizes for
all datasets and models. There is no discernible relationship between
the correlation and the feature sample size in the top row containing
datasets which are smaller in size but contain a large number of
features. A slight relationship can only be observed in the bottom
right subplots which contain datasets which are larger in size but
contain less number of features.

\highlight{\textbf{Answer to RQ3:} DFM and MFM convey similar
information as the training sample size changes but not when the
feature sample size changes.}
\section{Implications}\label{sec:implications}

\subsection{Data Drift}\label{sec:discuss-data-drift}

Results from RQ1 indicate that the DFM and MFM convey the same
information when the distribution and consequently the fairness
properties of the training data changes. ML systems running in a
production environment are often monitored to detect degradation in
model performance. As shown in Figure \ref{fig:ml-lifecycle}, a
standard practice is to combine the data encountered by the model in
the production environment with its predictions to create the training
set for the next iteration \cite{biessmann2021automated}. Since data
reflects the real world, change in its underlying distribution over
time is eminent. Our results indicate that DFM can be used as a early
warning system to identify fairness related data drifts in automated
ML pipelines.

\subsection{Fairness, data size and correctness trade-off}\label{sec:discuss-fair-eff-perf-trade}

\begin{figure}
  \centering
  \includegraphics[width=\linewidth]{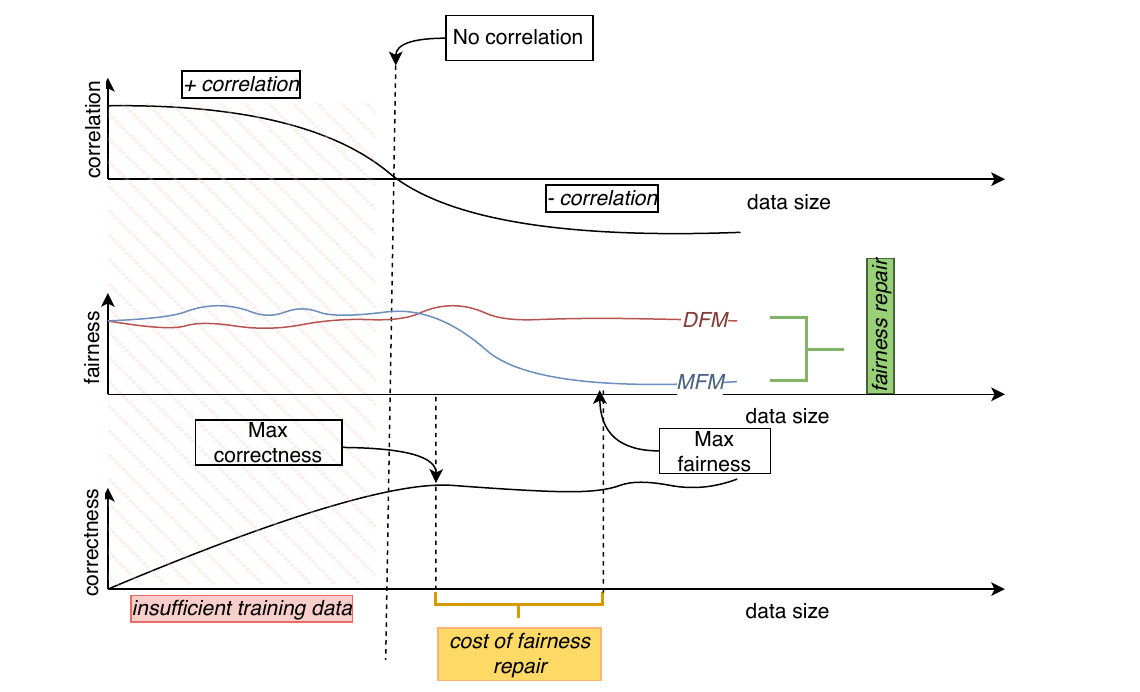}
  \caption{Visual representation of trade-offs between fairness, data
    size and correctness of ML models.}
  \label{fig:tradeoff}
\end{figure}

Results from RQ2 show that the quantity of training data significantly
impacts the relationship between DFM and MFM (explained visually in
Figure~\ref{fig:tradeoff}). With smaller training sizes, there's
a positive correlation between DFM and MFM, indicating bias in both
training data and model predictions. As training data increases, this
correlation diminishes, suggesting models learn to make fairer
predictions despite biases in training data. However, \emph{Zhang et
al. (2021)}~\cite{zhang2021ignorance} highlight that data quality is
also crucial; merely increasing data quantity doesn't necessarily
resolve model biases. The study also notes a trade-off between model
efficiency, performance, and fairness. Practitioners might reduce
training data for efficiency, potentially impacting model performance
and necessitating additional fairness mitigation efforts. Conversely,
larger training datasets can reduce bias but require more
computational resources. This balance between fairness, efficiency,
and performance is a key consideration in ML system development and
operation.

\subsection{Test Reduction}\label{sec:discuss-test-red}

Results from RQ3 show a positive correlation between DFM and MFM as
training sample size changes. This suggests that DFM can help
practitioners identify fairness issues early, potentially saving the
costs and energy associated with a full training cycle. Early
detection of bias with DFM might also indicate problems in data
collection or system design. However, this approach doesn't apply when
altering the feature sample size of the training set, as larger
feature samples usually enhance model fairness. Since no existing
fairness metrics consider feature influence at the data level, it's
advisable for ML practitioners to assess fairness both before and
after training when experimenting with feature sample size.



\section{Threats to Validity and Future Work}\label{sec:threats}

This study evaluates ML model fairness using two group fairness
metrics: Disparate Impact and Statistical Parity Difference due to the
absence of Python library implementations for other metrics like
Average Absolute Odds Difference and Equal Opportunity Difference. The
analysis reveals a need for more data-centric fairness metrics and
highlights limitations in current metrics, particularly their focus on
the size of the training set. To validate the results of the
correlation analysis, we additionally employ linear regression
analysis using ordinary least squares from the statsmodels library.
The approximation in simulating distribution changes with smaller
training sample sizes and the inconsistency in dataset sizes present
opportunities for improvement. While the correlation analysis is
promising, its practical application remains unclear and calls for
further research.

\section{Conclusion}

This study introduces a novel approach to ML fairness testing by
evaluating fairness both before and after training---using metrics to
quantify bias in training data and model predictions, respectively.
This ``data-centric'' approach is the first step towards integrating
fairness testing into the ML development lifecycle. The study
empirically analyzes the relationship between model-dependent and
independent fairness metrics, finding a linear relationship when
training data size and distribution change. The results suggest that
testing for fairness before training an ML model is a cost-effective
strategy for identifying fairness issues early in ML pipelines, and
can aid practitioners navigate the complex landscape of fairness
testing. As an extension of this study, we wish to evaluate the
effectiveness of data fairness metrics in real-world ML systems.
\bibliographystyle{ACM-Reference-Format}
\bibliography{report}

\end{document}